\documentclass[sigconf]{acmart}

\usepackage[utf8]{inputenc}
\usepackage[english]{babel}
\usepackage{hyperref}
\usepackage{graphicx}
\usepackage{numprint}
\usepackage{xcolor}
\usepackage{booktabs} 
\usepackage{microtype}
\usepackage{todonotes}

\usepackage{amssymb}
\usepackage{latexsym}
\usepackage{todonotes}
\usepackage{dcolumn}
\usepackage{graphics}
\usepackage{url}
\usepackage{longtable}
\usepackage{supertabular}
\usepackage{pdflscape}
\usepackage{pifont}

\usepackage{rotating}
\usepackage{tabularx}
\usepackage{lscape}
\usepackage{multirow}
\usepackage{multicol}
\usepackage{float}
\usepackage{footnote}
\usepackage{verbatim}
\usepackage{hyperref}
\usepackage{url}
\usepackage{mathtools}
\usepackage{xspace}
\usepackage{listingsutf8}
\usepackage{color}
\usepackage{tikz}
\usepackage{lipsum}
\usepackage[ruled,vlined]{algorithm2e}
\usepackage{mdframed}
\usepackage{wrapfig}
\usepackage[nolist]{acronym}
\usepackage{adjustbox}
\usepackage{array}

\newcommand{\furl}[1]{\footnote{\scriptsize \url{#1}}}

\newcolumntype{R}[2]{%
    >{\adjustbox{angle=#1,lap=\width-(#2)}\bgroup}%
    l%
    <{\egroup}%
}
\newcommand*\rot{\multicolumn{1}{R{90}{1em}}}

\newcommand{\tablefont}[1]{\fontsize{3mm}{3.2mm}\selectfont}

\begin{acronym}[UML]
	\acro{AOS}{Agricultural Ontology Services}
	\acro{AGRIS}{Agricultural Science and Technology}
	\acro{API}{Application Programming Interface}
	\acro{A2KB}{Annotation to Knowledge Base}
	\acro{BPSO}{Binary Particle-Swarm Optimization}
	\acro{BPMLOD}{Best Practices for Multilingual Linked Open Data}
	\acro{BFS}{Breadth-First-Search}
	\acro{CBD}{Concise Bounded Description}
	\acro{COG}{Content Oriented Guidelines}
	\acro{CSV}{Comma-Separated Values}
	\acro{CCR}{cross-document co-reference}
	\acro{DPSO}{Deterministic Particle-Swarm Optimization}
	\acro{DALY}{Disability Adjusted Life Year}
	\acro{D2KB}{Disambiguation to Knowledge Base}

	\acro{ER}{Entity Resolution}
	\acro{EM}{Expectation Maximization}
	\acro{EL}{Entity Linking}
	\acro{FAO}{Food and Agriculture Organization of the United Nations}
	\acro{GIS}{Geographic Information Systems}
	\acro{GHO}{Global Health Observatory}
	\acro{HDI}{Human Development Index}
	\acro{ICT}{Information and communication technologies}
    \acro{KB}{Knowledge Base}
	\acro{LR}  {Language Resource}
	\acro{LD}  {Linked Data}
	\acro{LLOD}  {Linguistic Linked Open Data}
	\acro{LIMES}{LInk discovery framework for MEtric Spaces}
	\acro{LS}  {Link Specifications}
	\acro{LDIF}{Linked Data Integration Framework}
	\acro{LGD} {LinkedGeoData}
	\acro{LOD} {Linked Open Data}
	\acro{MSE}{Mean Squared Error}
	\acro{MWE}{Multiword Expressions}
	\acro{NIF}{NLP Interchange Format}
	\acro{NIF4OGGD}{NLP Interchange Format for Open German Governmental Data}
	\acro{NLP}{Natural Language Processing}
	\acro{NER}{Named Entity Recognition}
	\acro{NED}{Named Entity Disambiguation}
	\acro{NEL}{Named Entity Linking}
	\acro{NN}{Neural Network}
	\acro{OSM}{OpenStreetMap}
	\acro{OWL}{Web Ontology Language}
	\acro{PFM}{Pseudo-F-Measures}
	\acro{PSO}{Particle-Swarm Optimization}
	\acro{QA}{Question Answering}
	\acro{RDF}{Resource Description Framework}
	\acro{SKOS}{Simple Knowledge Organization System}
	\acro{SPARQL}{SPARQL Protocol and RDF Query Language}
	\acro{SRL}{Statistical Relational Learning}
	\acro{SF}{surface forms}
	\acro{UML}{Unified Modeling Language}
	\acro{WHO}{World Health Organization}
	\acro{WKT}{Well-Known Text}
	\acro{W3C}{World Wide Web Consortium}
	\acro{YPLL}{Years of Potential Life Lost}
\end{acronym}  

\begin{document}

\title[MAG: Multilingual Entity Linking]{MAG: A Multilingual, Knowledge-base Agnostic and Deterministic Entity Linking Approach}


\author{Diego Moussallem}
\affiliation{
  \institution{AKSW Research Group}
  \institution{University of Leipzig}
  \country{Germany}}
\email{moussallem@informatik.uni-leipzig.de}

\author{Ricardo Usbeck}
\affiliation{
  \institution{Data Science Group}
  \institution{University of Paderborn}
  \country{Germany}}
\email{ricardo.usbeck@upb.de}

\author{Michael R{\"o}der}
\affiliation{
  \institution{Data Science Group}
  \institution{University of Paderborn}
  \country{Germany}}
\email{michael.roeder@upb.de}

\author{Axel-Cyrille Ngonga Ngomo}
\affiliation{
  \institution{Data Science Group}
  \institution{University of Paderborn}
  \country{Germany}}
\email{axel.ngonga@upb.de}

\renewcommand{\shortauthors}{Moussallem et al.}

\begin{abstract}
Entity linking has recently been the subject of a significant body of research. Currently, the best performing approaches rely on trained mono-lingual models. Porting these approaches to other languages is consequently a difficult endeavor as it requires corresponding training data and retraining of the models. We address this drawback by presenting a novel multilingual, knowledge-base agnostic and deterministic approach to entity linking, dubbed MAG. MAG is based on a combination of context-based retrieval on structured knowledge bases and graph algorithms. We evaluate MAG on 23 data sets and in 7 languages. Our results show that the best approach trained on English datasets (PBOH) achieves a micro F-measure that is up to 4 times worse on datasets in other languages. MAG on the other hand achieves state-of-the-art performance on English datasets and reaches a micro F-measure that is up to 0.6 higher than that of PBOH on non-English languages.
\end{abstract}

%
%


\copyrightyear{2017}
\acmYear{2017}
\setcopyright{acmlicensed}
\acmConference{K-CAP 2017: Knowledge
Capture Conference}{December 4--6, 2017}{Austin, TX,
USA}
\acmPrice{15.00}
\acmDOI{10.1145/3148011.3148024}
\acmISBN{978-1-4503-5553-7/17/12}

\ccsdesc[500]{Information systems~Information extraction}
\ccsdesc[500]{Computing methodologies~Natural language processing}

\keywords{Entity Linking; Multilingual; Named Entity Disambiguation}


\maketitle

\section{Introduction}
\label{sec:introduction}

More than one exabyte of data is added to the Web every day.\footnote{\url{https://tinyurl.com/kx6baxx}}  
Automatic extraction of knowledge from this data demands the use of efficient \ac{NLP} techniques such as text aggregation, text summarization and knowledge extraction.  
One of the most important \ac{NLP} tasks is \ac{EL}, also known as \ac{NED}. The goal here is as follows: Given a piece of text, a reference knowledge base $K$ and a set of entity mentions in that text, map each entity mention to the corresponding resource in $K$.   

Several challenges have to be addressed when dealing with \ac{EL}. For example, an entity can have a large number of \ac{SF} (also known as labels) due to synonymy, acronyms and typos. For example, \texttt{New York City}, \texttt{NY} and \texttt{Big Apple} are all labels for the same entity. Moreover, multiple entities can share the same name due to homonymy and ambiguity. For example, both the state and the city of New York are called \texttt{New York}. Despite the complexity of the endeavor, \ac{EL} approaches have achieved increasingly better results over the past few years by relying on trained machine learning models (see \cite{gerbil} for an overview). A portion of these approaches claim to be multilingual and rely on using cross-lingual dictionaries. However, our experiments (see Section \ref{sec:evaluation}) show that the underlying models being trained on English corpora make them prone to failure when migrated to a different language. For example, while PBOH~\cite{PBOH} achieves an average micro F-measure of 0.69 on corpora in English, it only achieves an average micro F-measure of 0.31 on other languages.

We alleviate this problem by presenting MAG, a novel multilingual \ac{EL} approach. MAG (Multilingual AGDISTIS) is based on concepts similar to those underlying AGDISTIS \cite{AGDISTIS_ISWC} but goes beyond this approach  
by relying on time-efficient graph algorithms combined with language-independent features to link entities to a given reference knowledge base. Hence, MAG is knowledge-base agnostic, i.e., it can be deployed on any reference knowledge base $K$. Our approach is also deterministic and does not rely on any trained model. Hence, it can be deployed on virtually any language.

The main contributions of this paper can be summarized as follows:
\begin{itemize}
    \item We present a novel multilingual and deterministic approach for \ac{EL} which combines lightweight and easily extensible 
    graph-based algorithms with a new context-based retrieval method.
    \item MAG features an innovative candidate generation method which relies on various filter methods and search types for a better candidate selection.
    \item We provide a thorough evaluation of our overall system on 23 data sets using the GERBIL platform~\cite{gerbil}. Our results show that MAG achieves state-of-the-art performance on English. In addition, MAG outperforms all state-of-the-art approaches on 6 non-English data sets.
\end{itemize}

MAG was implemented within the AGDISTIS framework.\footnote{\url{https://github.com/AKSW/AGDISTIS}} The version of MAG used in this paper and also all experimental results are publicly available\footnote{\url{https://github.com/AKSW/AGDISTIS/releases/tag/v1.0-mag}}\footnote{\url{http://faturl.com/magexp/?open&selected=0}}.

\section{Related Work}
\label{sec:relatedwork}
\ac{EL} approaches can be subdivided into 2 different classes: 1) English-only approaches which are based and evaluated on English data sets and 2) multilingual approaches.

\textbf{English-only} 
In 2013, Van Erp et al.~\cite{vanErp2013} proposed an approach, dubbed NERD-ML, for entity recognition tailored for extracting entities from tweets. 
This approach relies on entity type classification over a rich feature vector composed of a set of linguistic components. 
Next to that, a plethora of other approaches including ~\cite{Cucerzan07,milne2008learning,rat:rot,han2011collective,chisholm2015entity,luo2015joint,francis2016capturing,zhangcontext2016} have been developed. These approaches mostly rely on graph algorithms and/or on machine-learning techniques. However, most of them do not offer a webservice nor an implementation that is publicly available and are thus difficult to compare.

\textbf{Multilingual} 
A large number of multilingual approaches have been developed over the years. DBpedia Spotlight~\cite{spotlight} combines \ac{NER} and \ac{NED} models based on a vector-space representation of entities and the use of cosine similarity for performing the disambiguation task. Hoffart et al.~\cite{AIDA} present AIDA which is based on the YAGO2 \ac{KB}. Gad-Elrab et al.~\cite{gad2015named} have implemented AIDA for other languages. KEA~\cite{Steinmetz2013} is based on a fine-granular context model that takes into account heterogeneous text sources as well as text created by automated multimedia analysis. 
Dojchinovski et al.~\cite{Dojchinovski:2013:ECMLPKDD13} presents Entity-classifier.eu which is based on hypernyms and a Wiki\-pedia-based entity classification system which identifies salient words. The input is transformed to a lower dimensional representation keeping the same quality of output for all sizes of input text. In 2014, Zhang et al.~\cite{xlisa} presented x-Lisa, a three-step pipeline based on cross-lingual Linked Data lexica that harnesses the multilingual Wikipedia. 
Navigli et al.~\cite{babelfy} proposed Babelfy, which is based on random walks and a densest subgraph algorithm 
and relies on the BabelNet semantic network~\cite{NavigliPonzetto:12aij} as background knowledge. Usbeck et al.~\cite{AGDISTIS_ISWC} presented AGDISTIS, a KB-agnostic entity disambiguation approach based on string similarity measures and the graph-based HITS algorithm. WAT~\cite{piccinno2014tagme} is the successor of TagME~\cite{TagMe2} which includes a re-design of all TagME components, namely, the spotter, the disambiguator, and the pruner.
In 2015, Consoli and Recupero~\cite{fred_typing} presented FRED, a novel machine reader which extends TagMe with entity typing capabilities. Zwicklbauer et al.~\cite{doser} presented DoSer, an approach akin to AGDISTIS but based on entity embeddings. 
Later, Ganea et al.~\cite{PBOH} introduced PBOH, 
a probabilistic graphical model which uses pairwise Markov Random Fields. 
Our related work study suggests that only three of the related approaches are multilingual and deterministic  without domain restriction. 
These are VINCULUM~\cite{ling2015design}, AGDISTIS~\cite{AGDISTIS_ISWC} and QVC~\cite{wang2015language}. 
However, VINCULUM, QVC and other recent supervised approaches~\cite{sil2016one,yahoo} could not be included in our evaluation because neither their code nor a webservice were made available publicly, and the approaches could not be reconstructed  based on the respective papers only. 

\section{The MAG Approach}
\label{sec:approach}
\begin{figure*}
\centering
\includegraphics[scale=0.50]{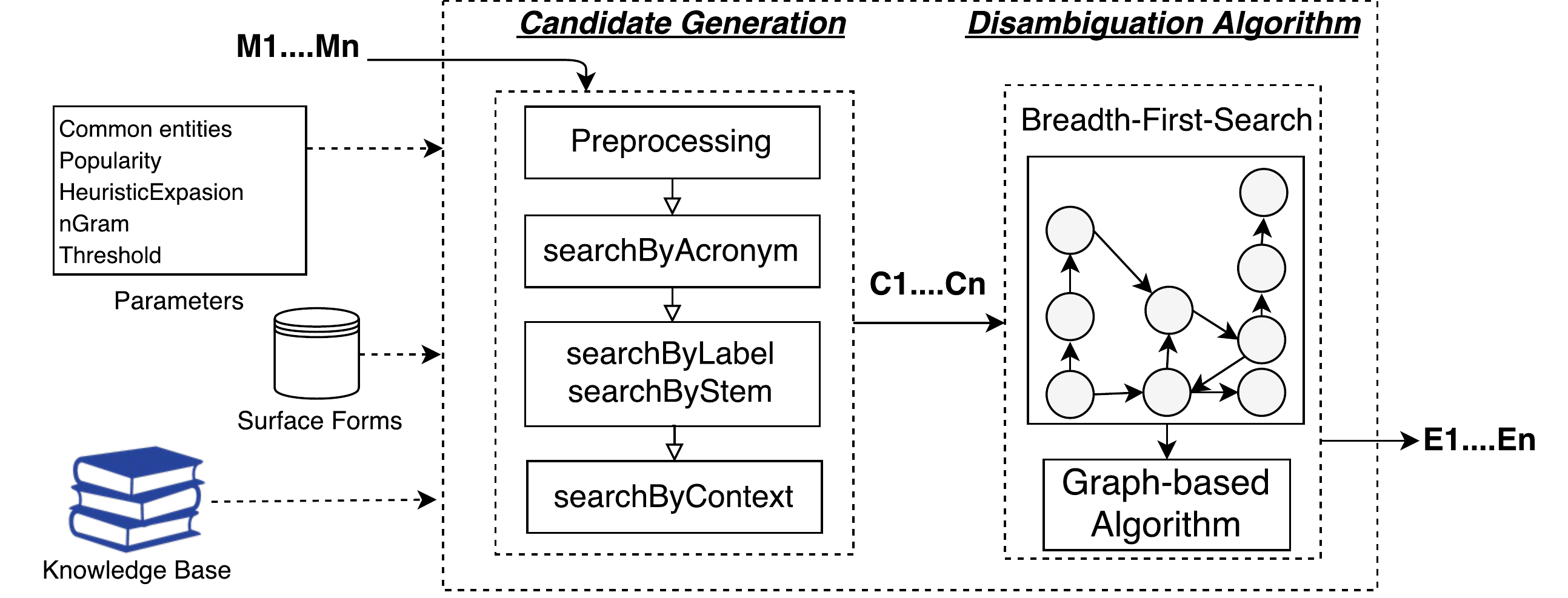}
\caption{MAG architecture overview.}
\label{fig:architecture}
\end{figure*}

In this section, we present MAG in detail. Throughout this work, we rely on the following formal definition of \ac{EL}.  
\begin{definition}
Entity Linking:
Let $\mathcal{E}$ be a set of entities from a \ac{KB} 
and 
$\mathcal{D}$ be a document containing potential mentions of entities \textbf{m} = $({m}_1,\dots,{m}_n)$.
The goal of an entity linking system is to generate an assignment $\mathcal{F}$ of mentions to entities with $\mathcal{F}(\textbf{m}) \in  (\mathcal{E \cup \{\epsilon\})}^n$ for the document $\mathcal{D}$, where $\epsilon$ stands for an entity that is not in the \ac{KB}. 
\end{definition}
The \ac{EL} process implemented by MAG consists of two phases. Several indexes are generated during the offline phase. 
The entity linking per se is carried out during the online phase and consists of two steps: 1) candidate generation and 2) disambiguation.
An overview can be found in~\autoref{fig:architecture}.

\subsection{Offline Index Creation}
\label{sec:index}
MAG relies on the following five indexes: surface forms, person names, rare references, acronyms and context.

\textbf{Surface forms.}
MAG relies exclusively on structured data to generate surface forms for entities so as to remain KB-agnostic.  
%
%
%
%
For each entity in the reference \ac{KB}, our approach harvests all labels of the said entity as well as its type and indexes them.\footnote{In our implementation, we relied on predicates such as \texttt{rdfs:label} and \texttt{rdf:type}.} 
Additional \ac{SF}s can be collected from different sources~\cite{AGDISTIS_ISWC,bryl2015gathering}.

\textbf{Person names} - This index accounts for the variations in names for referencing persons~\cite{krahmer2012computational} across languages and domains. Persons are referred to by different portions of their names. For example, the artist \texttt{Beyonc\'e Giselle Knowles-Carter} is often referred to as \texttt{Beyonc\'e} or \texttt{Beyonc\'e Knowles}. In Brazil, she is also known as \texttt{Beyonc\'e Carter} and \texttt{Beyonc\'e G. Knowles}. Moreover, languages such as Chinese and Japanese put the family name in front of the given name (in contrast to English, where names are written in the reverse order).
We address the problem of labelling persons by generating all possible permutations of the words within the known labels of persons and adding them to the index of names.

\textbf{Rare references} - This index is created if textual descriptions are available for the resources of interest (e.g., if resources have a \texttt{rdfs:comment} property). A large number of textual entity descriptions provide type information pertaining to the resource at hand, as in the example ``Michael Joseph Jackson was an American singer ..."\footnote{See \texttt{rdfs:comment} of \url{http://dbpedia.org/resource/Michael_Jackson}.}. Hence, we use a POS tagger (the Stanford POS tagger~\cite{toutanova2000enriching} in our implementation) on the first line of a resource's description and collect any noun phrase that contains an adjective. For example, we can extract the supplementary \ac{SF} \texttt{American singer} for our example. This is similar to~\cite{CETUS_2015}. 

\textbf{Acronyms} - Acronyms are used across a large number of domains, e.g., in news (see AIDA and MSNBC data sets).
We thus reuse a handcrafted index from \-STANDS4.\footnote{See \url{http://www.abbreviations.com/}} 

\textbf{Context} - Previous works\cite{doser} rely on semantic embeddings such as Word\-2\-vec~\cite{mikolov2013distributed} to create indexes that model the words  surrounding resource mentions in textual corpora. 
Given that we aim to be KB-agnostic and deterministic, our context index relies on the \ac{CBD}\footnote{\url{https://www.w3.org/Submission/CBD/}} of resources. The literals found in the \ac{CBD} of each resource are first freed of stop words. Then, each preprocessed string is added as an entry that maps to the said resource.

All necessary information for recreating our indexes or building new indexes for other \ac{KB}s can be found in our Wiki.\footnote{\url{https://github.com/AKSW/AGDISTIS/wiki/3-Running-the-webservice}}

\subsection{Candidates Generation}
\label{sec:candidates}
The candidate generation and the disambiguation steps occur online, i.e., when MAG is given a document and a set of mentions to disambiguate. The goal of the candidate generation step is to retrieve a tractable number of candidates for each of the mentions. These candidates are later inserted into the disambiguation graph, which is used to determine the mapping between entities and mentions (see Section \ref{sec:disambiguation}).  
    
First, we \textbf{preprocess mentions} to improve the retrieval quality. Before using common normalization techniques, we apply a filter for separating acronyms. The acronym filter detects acronyms by looking for strings made up of 5 uppercase letters or less.
For example, ``PSG" is the acronym of ``Paris Saint-Germain". 
In the case where a mention is considered an acronym, all further preprocessing steps are skipped. Otherwise, we use basic normalization methods which remove punctuation, symbols and additional white spaces. We use regular expressions to normalize the structure of strings. 
For example, our procedure will map strings to the lower case except for the first letter. Therewith, ``NEW YORK'' is normalized to ``New York''. Furthermore, our preprocessing is capable of recognizing mentions such as camel cases ``AmyWinehouse" and adds a space between lower-case and upper-case letters (i.e. true casing technique~\cite{lita2003truecasing}).

 


The second step of the candidate generation, the \textbf{candidate search}, is divided into three parts:

\textbf{By Acronym} - If a mention is considered an acronym by our preprocessing, we expand the mention with the list of possible names from the acronym index mentioned above. For example, ``PSG" is replaced by ``Paris Saint-Germain".

\textbf{By Label} -  This search relies on our \ac{SF} index. First, MAG retrieves candidates for a mention using exact matches to their respective principal reference. For example, the mention ``Barack Obama" and the principal reference of the former president of the USA, which is also ``Barack Obama", match exactly. In cases where we find a string similarity match with the main reference of 1.0, the remaining steps are skipped. If this search does not return any candidates, MAG starts a new search using a trigram similarity threshold $\sigma$ over the \ac{SF} index. In cases where the set of candidates is still empty, MAG stems the mention and repeats the search.
For example, MAG stems ``Northern India" to ``North India" to account for linguistic variability. The stemming process is not initialized in the preprocessing step since it is a technique which can induce additional errors~\cite{singh2016text}. For both search types, i.e. search by acronyms and labels, we apply trigram similarity for retrieving possible candidates.

\begin{figure}
\centering
\includegraphics[scale=0.4]{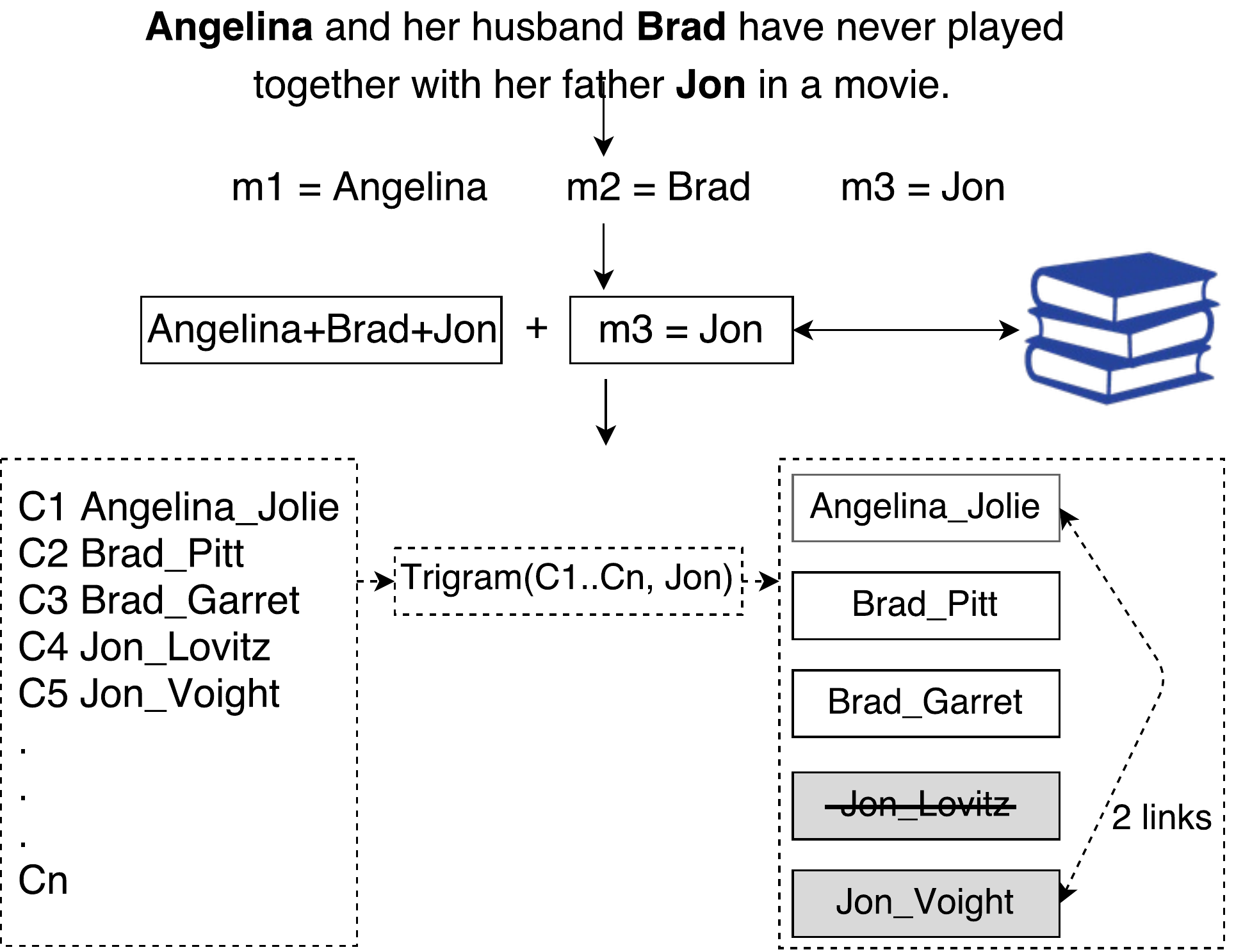}
\caption{Search using the context index. White boxes on the right side depict candidates discarded by the trigram filter.}
\label{fig:context}
\end{figure}

\textbf{By Context} - Here, we use two \textbf{post-search filters} to find possible candidates from the context index. 
Before applying both filters, MAG extracts all entities contained in the input document.
These entities are used as an addition while searching a mention in the context index.
This search relies on TF-IDF~\cite{ramos2003using} which reflects the importance of a word or string in a document corpus relative to the relevance in its index.
Afterwards, MAG first filters unlikely candidates by applying trigram similarity. Second, MAG retrieves all direct links among the remaining candidates in the \ac{KB}.
Our approach uses the number of connections to find highly related entity sets for a specific mention. 
This is similar to finding a dense subgraph~\cite{AIDA}.
\autoref{fig:context} illustrates an example which contains three ambiguous entities, namely ``Angelina", ``Brad" and ``Jon". Regarding the mention ``Jon", MAG searches the context index using ``[(Angelina + Brad + Jon) + Jon]" as a query. MAG keeps only ``Jon\_Lovitz" and ``Jon\_Voight" after trigram filtering. Only ``Jon\_Voight", the father of ``Angelina\_Jolie", has direct connections with the other candidates and is thus chosen.

To improve the quality of candidates, the popularity of a given entity is a good ranking factor. If MAG's configuration uses this factor, the number of candidates retrieved from the index is increased and then the result is sorted. After this sorting step, MAG returns the top 100 candidates. The popularity is calculated using Page Rank~\cite{page1999pagerank} over the underlying \ac{KB}. {If we are unable to leverage Page Rank on certain KB's, we fall back to a heuristic of inlinks and outlinks.} MAG's candidate generation process is shown in Algorithm~\ref{alg:findingCandidates}.\footnote{The in-document, co-reference resolution is based on earlier works.~\cite{AGDISTIS_ISWC}}

\begin{algorithm}[htb]
\small
\KwData{Mention $m_i$, $\sigma$ trigram similarity threshold}
\KwResult{$C$ candidates found}
$C, \bar C \longleftarrow \emptyset$\;
{\bf mention } $\longleftarrow$ {\bf Preprocessing($m$)}; {\bf mention } $\longleftarrow$ {\bf Co-reference($m$)}\;
\If{{\bf containsAcronym(m)}} {
                $ \displaystyle \bar C \longleftarrow$ {\bf searchByAcronym($m$)}\;}
\Else{
                $ \displaystyle \bar C \longleftarrow$ {\bf searchByLabel($m$)}\;
                }
\For{{\bf c} $\in \bar C$}{
    \If{$\neg${\bf c .matches([0-9]$^+$)}}{
            \If{{\bf trigramSimilarity(c, $m$)} $ = 1.0$}{
                $C \longleftarrow C \cup $ {\bf c; return\;
            }
        }
        \ElseIf{{\bf trigramSimilarity(c, $m$)} $ \geq \sigma$}{
                $C \longleftarrow C \cup $ {\bf c\;
            }
        }
    }
}
                $ \displaystyle \bar C \longleftarrow$ {\bf searchByContext($m$)}\;
                \For{{\bf c} $\in \bar C$}{
        \If{{\bf trigramSimilarity(c, $m$)}$ \geq \sigma$}{
\For{{\bf c} $\in \bar C$}{
    \If{{\bf c.directLinks($\bar C$)} $\ge 0$}{
                $C \longleftarrow C \cup $ {\bf c}\;    
    }
}
        }
    }
\caption{Candidates Generation.}
\label{alg:findingCandidates}
\end{algorithm}

\subsection{Entity Disambiguation Algorithm}
\label{sec:disambiguation}

After the candidate generation step, the computation of the optimal candidate to mention assignment starts by constructing a disambiguation graph $G_d$ with depth $d$ similar to the approach of AGDISTIS.
\begin{definition}
Knowledge Base: We define \ac{KB} $K$ as a directed graph $G_K = (V, E)$ where the nodes $V$ are resources of $K$, the edges $E$ are properties of $K$ and $x,y\in V, (x,y) \in E \Leftrightarrow \exists p : (x, p, y) \mbox{ is a triple in }K$.
\end{definition}
Given the set of candidates $C$, we begin by building an initial graph $G_0 = (V_0, E_0)$ where $V_0$ is the set of all resources in $C$ and $E_0=\emptyset$. Starting with $G_0$ the algorithm expands the graph using \ac{BFS} technique in order to find hidden paths among candidates. The extension of a graph is $G_i = (V_i, E_i)$ to a graph $\rho(G_i) = G_{i+1} = (V_{i+1}, E_{i+1})$ with $i=0, \ldots, d$.
The $\rho$ (\ac{BFS}) operator iterates $d$ times on the input graph $G_0$ to compute the initial disambiguation graph $G_d$. 
After constructing $G_d$, we need to identify the correct candidate node for a given mention. Here, we rely on HITS~\cite{HITS} or Page Rank~\cite{page1999pagerank} as disambiguation graph algorithms. This choice comes from a comparative study of the differences between both~\cite{devi2014comparative}.

\textbf{HITS} uses hub and authority scores to define a recursive relationship between nodes. An authority node is a node that many hubs link to and a hub is a node that links to many authorities. The authority values are equal to the sum of the hub scores of each node that points to it. The hub values are equal to the sum of the authority scores of each node that it points to. According to previous work~\cite{AGDISTIS_ISWC}, we chose 20 iterations for HITS which suffice to achieve convergence in general.

\textbf{Page Rank} has a wide range of implementations. We implemented the general version in accordance with~\cite{page1999pagerank}. Thus, we defined the possibility of jumping from any node to any other node in the graph during the random walk with a probability $\alpha = (1-w) = 0.15$. We empirically chose 50 Page Rank iterations which has shown to be a reasonable number for \ac{EL}~\cite{doser}. We assigned a standard weight $w = 0.85$ for each node. Finally, the sum is calculated by spreading the current weight divided by outgoing edges.

Independent of the chosen graph algorithm, the highest candidate score among the set of candidates $C$ is chosen as correct disambiguation for a given mention $m_i$.
The entire process is presented in Algorithm~\ref{algooverview}.
Note, MAG also considers {emergent entities}~\cite{Hoffart:2014:DEE:2566486.2568003} and assigns a new URI to them.\footnote{\url{https://www.w3.org/TR/cooluris/}}
\begin{definition}
Emergent entity: If the candidate generation step fails to retrieve any candidate from the target \ac{KB}, we assume the mention belongs to an emergent entity.
\end{definition}

\begin{algorithm}[htb]
\small
\caption{Disambiguation Algorithm based on HITS and Page Rank.}\label{algooverview}
\KwData{$M=\{m_1,m_2\dots m_n\}$ mentions, $d$ depth, $k$ number of iterations}
\KwResult{$C = \{c_1,c_2\dots c_n\}$ identified candidates for named entities}
$E \longleftarrow \emptyset$\;
$V \longleftarrow${\bf insertCandidates($N, \sigma$)}\;
$G \longleftarrow (V,E)$\;
$G \longleftarrow${\bf
breadthFirstSearch($G,d$)}\;
\If{{\bf HITS}} {
                {\bf HITS($G(V,E), k$)}\;
{\bf sortAccordingToAuthorityValue(V)}\;
\For{$m_i \in M$} {
    \For{$v \in V$}{
        \If{$v$ {\bf is a candidate for} $m_i$  }{
              {\bf store($m_i$,$v$)}\;
              {\bf break}\;
          }
     }
}}
\If{{\bf Page Rank}} {
                {\bf Page Rank($G(V,E), k, \alpha$)}\;
{\bf sortSumValue(V)}\;
\For{$m_i \in M$} {
    \For{$v \in V$}{
        \If{$v$ {\bf is a candidate for} $m_i$  }{
              {\bf store($m_i$,$v$)}\;
              {\bf break}\;
          }
     }
}}
\end{algorithm}


\section{Evaluation}
\label{sec:evaluation}
\subsection{Goals}
The aim of our evaluation is three-fold. 
First, we aim to measure the performance of MAG on 17 data sets and compare it to the state of the art for EL in English.  
Second, we evaluate MAG's portability to other languages. To this end, we compare MAG and the multilingual state of the art using 6 data sets from different languages. For both evaluations we use HITS and Page Rank.
Third, we carry out a fine-grained evaluation providing a deep analysis of MAG using the method proposed in~\cite{waitelonis2016don}. 
Throughout our experiments, we used DBpedia as reference \ac{KB}.

\subsection{Experimental setup}
\label{sec:data sets}
For our evaluation, we rely on the GERBIL platform \cite{gerbil} focusing on the \textbf{Disambiguation to KB (D2KB)} experiment type. The task is to map a set of given mentions to entities from a given \ac{KB} or to $\epsilon$ (meaning the resource cannot be found in the reference \ac{KB}). 

All data sets (see ~\autoref{tab:corpus_stats} for an overview) are integrated into GERBIL for the sake of comparability.\footnote{The TAC-KBP data sets could not be included in our evaluation because of their redistribution license.} \textbf{ACE2004} originates from~\cite{rat:rot} and is a subset of the ACE co-reference documents. The annotations were obtained through crowd-sourcing where annotators linked the first mention of each reference to Wikipedia. 
\textbf{AIDA/CoNLL} is divided into 3 chunks: Training, TestA and TestB and exclusively contains annotations based on named entities. This manually annotated data set was used to evaluate AIDA, and stems from the CoNLL 2003 shared task~\cite{conll2003}.
\textbf{AQUAINT} contains annotations of the first mention of each entity in its news-wire documents~\cite{milne2008learning}. 
\textbf{Spotlight} was released along with DBpedia Spotlight~\cite{spotlight}. The manually annotated data set contains short texts of named entities and common entities such as \texttt{cancer} and \texttt{home}.
\textbf{IITB} was created in 2009 and has the highest en\-ti\-ty/do\-cu\-ment-den\-si\-ty of all corpora~\cite{kulkarni2009collective}.
\textbf{KORE50}'s~ aim is to stress test EL systems through difficult disambiguation tasks using highly ambiguous mentions using hand-crafted sentences.\footnote{\url{http://www.yovisto.com/labs/ner-benchmarks/}}
The \textbf{Microposts 2014} data sets were created for the "Making Sense of Microposts" challenge and contains only tweets. 
\textbf{MSNBC} was introduced in 2007 by~\cite{Cucerzan07}. The data set contains news documents with rare \ac{SF} and a distinctive lexicalization.
\textbf{N$^3$ Reuters-128} comprises 128 news articles  which were sampled from the Reuters-21578 news articles data set randomly and annotated manually by domain experts.
\textbf{N$^3$ RSS-500} consists of data scrapped from 1,457 RSS feeds~\cite{GER+13}. The list includes all major worldwide newspapers and a wide range of topics. The corpus was annotated manually by domain experts.
\textbf{OKE 2015} was used in the OKE challenge~\cite{okechallenge}. The data sets were curated manually and are divided into 3 subsets. 
\textbf{N$^3$ news.de} is a real-world data set collected from 2009 to 2011. It contains documents from the German news portal \url{news.de}. 
\textbf{DBpedia Abstracts} is a large, multilingual corpus generated from enriched Wikipedia data of annotated Wikipedia abstracts from six languages~\cite{BrummerDH16}.\footnote{We reduced our test set to the first subset of provided abstracts for each language due to evaluation platform limits and display their characteristics in \autoref{tab:corpus_stats}.}

\begin{table}[!htb]
\centering
\caption{Data set statistics.}
\small 
\setlength\tabcolsep{1pt}
\begin{tabular}{@{}lcccc@{}}
\toprule
\textbf{Corpus} & \textbf{Language} & \textbf{Topic} & \textbf{Documents} & \textbf{Entities} \\
\midrule
ACE2004         & English & news      &   57 &   253\\ 
AIDA/CoNLL-Complete      & English & news      & 1393 &  34929\\
AIDA/CoNLL-Test A      & English & news      & 216 &  5917\\
AIDA/CoNLL-Test B      & English & news      & 231 &  5616\\
AIDA/CoNLL-Training      & English & news      &  946 &  23396\\
AQUAINT         & English & news      &   50 &  747\\ 
Spotlight Corpus & English & news &    58 &  330\\ 
IITB             & English & mixed     &  103 & 18308\\
KORE 50          & English & mixed     &   50 &   144\\
Microposts2014-Test   & English & tweets    & 1055 &   1256\\
Microposts2014-Train   & English & tweets    & 3395 &   3822\\
MSNBC            & English & news      &   20 &  747\\ 
N$^3$ Reuters-128 & English & news      &  128 &   880\\ 
N$^3$ RSS-500    & English & mixed & 500 &   1000\\  
OKE 2015 Task 1 evaluation set  & English & mixed     &  101 &   664\\
OKE 2015 Task 1 example set  & English & mixed     &  3 &   6\\
OKE 2015 Task 1 training set  & English & mixed     &  96 &   338\\
\midrule
N$^3$ news.de    & German  & news & 53 &   627\\
Dutch Abstract & Dutch & mixed & 39,300 & 385,259 \\
French Abstract & French & mixed & 38,197 & 346,448 \\
Spanish Abstract & Spanish & mixed & 37,663 & 452,628 \\
Italian Abstract & Italian & mixed & 36,432 & 310,775 \\
Japanese Abstract & Japanese & mixed & 38,823 & 316,982 \\
\bottomrule
\end{tabular}
\label{tab:corpus_stats}
\end{table}

\subsubsection{Results on English data sets}
\label{sec:english}

For this evaluation, we configured MAG to disambiguate named entities as well as common entities, to use the acronyms index and the popularity scores. According to ~\cite{AGDISTIS_ISWC}, N-gram at letter level achieved best results when set to 3. The $\sigma$ trigram threshold was optimal at 0.87\footnote{We chose the trigram threshold in accordance with the work of Usbeck. et~al.~\cite{AGDISTIS_ISWC} \url{http://titan.informatik.uni-leipzig.de/rusbeck/agdistis/appendix.pdf}} and  varying the depth of \ac{BFS} yields an optimal choice of 2. Therefore, we tuned the thresholds to $d=2$ and $\sigma=0.87$ for performing the evaluation of MAG. 
The English results are shown in the first part of \autoref{engfmeasure}.\footnote{\url{http://gerbil.aksw.org/gerbil/experiment?id=201701230012} and \url{http://gerbil.aksw.org/gerbil/experiment?id=201701260017}}
 
\begin{table*}[htb]
\footnotesize
\setlength\tabcolsep{3pt} 
\centering
\caption{Micro F-measure across approaches. Red entries are the top scores while blue represents the second best scores.}
\label{engfmeasure}
\begin{tabular}{@{}l|lclllllllllll|l|l@{}}
\rot{\textbf{Language}} & \textbf{Tools/Data sets}  & \rot{\textbf{AGDISTS}} & \rot{\textbf{AIDA}} & \rot{\textbf{Babelfy}} & \rot{\textbf{DBpedia}} & \rot{\textbf{DoSer}} & \rot{\textbf{entityclassifier.eu}} & \rot{\textbf{FRED}} & \rot{\textbf{Kea}} & \rot{\textbf{NERD-ML}} & \rot{\textbf{PBOH}} & \rot{\textbf{WAT}} & \rot{\textbf{xLisa}} & \rot{\textbf{MAG + HITS}} & \rot{\textbf{MAG + PR}} \\ \toprule
\multirow{17}{*}{\rotatebox{90}{English}}

& ACE2004 & 0.65 & 0.70 & 0.53 & 0.48 & {\color[HTML]{FE0000} \textbf{0.75}} & 0.50 & 0.00 & 0.66 & 0.58 & {\color[HTML]{2033FF} \textbf{0.72}} & 0.66 & 0.70 & 0.69 & 0.60 \\ 

&AIDA/CoNLL-Complete & 0.55 & 0.68 & 0.66 & 0.50 & 0.69 & 0.50 & 0.00 & 0.61 & 0.20 & {\color[HTML]{FE0000} \textbf{0.75}} & {\color[HTML]{2033FF} \textbf{0.71}} & 0.48 & 0.59 & 0.54\\ 

&AIDA/CoNLL-Test A & 0.54 & 0.67 & 0.65 & 0.48 & 0.69 & 0.48 & 0.00 & 0.61 & 0.00 & {\color[HTML]{FE0000} \textbf{0.75}} & {\color[HTML]{2033FF} \textbf{0.7}} & 0.45 & 0.59 & 0.54\\ 

&AIDA/CoNLL-Test B & 0.52 & 0.69 & 0.68 & 0.52 & 0.69 & 0.48 & 0.00 & 0.61 & 0.00 & {\color[HTML]{FE0000} \textbf{0.75}} & {\color[HTML]{2033FF} \textbf{0.72}} & 0.47 & 0.57 & 0.52\\ 

&{AIDA/CoNLL-Training} & 0.55 & 0.69 & 0.66 & 0.50 & 0.69 & 0.52 & 0.00 & 0.61 & 0.28 & {\color[HTML]{FE0000} \textbf{0.75}} & {\color[HTML]{2033FF} \textbf{0.71}} & 0.48 & 0.60 & 0.55\\ 

&{AQUAINT} & 0.52 & 0.55 & 0.68 & 0.53 & {\color[HTML]{FE0000} \textbf{0.82}} & 0.41 & 0.00 & 0.78 & 0.60 & {\color[HTML]{2033FF} \textbf{0.81}} & 0.73 & 0.76 & 0.67 & 0.68\\ 

&{Spotlight} & 0.27 & 0.25 & 0.52 & 0.71 & {\color[HTML]{FE0000} \textbf{0.81}} & 0.25 & 0.04 & 0.74 & 0.56 & {\color[HTML]{2033FF} \textbf{0.79}} & 0.67 & 0.71 & 0.65 & 0.66\\ 

&{IITB} & 0.47 & 0.18 & 0.37 & 0.30 & 0.43 & 0.14 & 0.00 & {\color[HTML]{2033FF} \textbf{0.48}} & 0.43 & 0.38 & 0.41 & 0.27 & {\color[HTML]{FE0000} \textbf{0.52}} & 0.43\\ 

&{KORE50} & 0.27 & {\color[HTML]{2033FF} \textbf{0.70}} & {\color[HTML]{FE0000} \textbf{0.74}} & 0.46 & 0.52 & 0.30 & 0.06 & 0.60 & 0.31 & 0.63 & 0.62 & 0.51 & 0.24 & 0.24\\ 

&{MSNBC} & 0.73 & 0.69 & 0.71 & 0.42 & {\color[HTML]{FE0000} \textbf{0.83}} & 0.51 & 0.00 & 0.78 & 0.62 & {\color[HTML]{2033FF} \textbf{0.82}} & 0.73 & 0.5 & 0.79 & 0.75\\ 

&{Microposts2014-Test} & 0.33 & 0.42 & 0.48 & 0.50 & {\color[HTML]{FE0000} \textbf{0.76}} & 0.41 & 0.05 & 0.64 & 0.52 & {\color[HTML]{2033FF} \textbf{0.73}} & 0.60 & 0.55 & 0.45 & 0.44\\ 

&{Microposts2014-Train} & 0.42 & 0.51 & 0.51 & 0.48 & {\color[HTML]{FE0000} \textbf{0.77}} & 0.00 & 0.31 & 0.65 & 0.52 & {\color[HTML]{2033FF} \textbf{0.71}} & 0.63 & 0.59 & 0.49 & 0.44\\ 

&{N3-RSS-500} & 0.66 & 0.45 & 0.44 & 0.20 & 0.48 & 0.00 & 0.00 & 0.44 & 0.38 & 0.53 & 0.44 & 0.45 & {\color[HTML]{FE0000} \textbf{0.69}} & {\color[HTML]{2033FF} \textbf{0.67}}\\ 

&{N3-Reuters-128} & 0.61 & 0.47 & 0.45 & 0.33 & {\color[HTML]{FE0000} \textbf{0.69}} & 0.00 & 0.41 & 0.51 & 0.41 & {\color[HTML]{2033FF} \textbf{0.65}} & 0.52 & 0.39 & {\color[HTML]{FE0000} \textbf{0.69}} & 0.64\\ 

&{OKE 2015 Task 1 evaluation set} & 0.59 & 0.56 & 0.59 & 0.31 & 0.59 & 0.00 & 0.46 & {\color[HTML]{FE0000} \textbf{0.63}} & 0.61 & {\color[HTML]{FE0000} \textbf{0.63}} & 0.57 & {\color[HTML]{2033FF} \textbf{0.62}} & 0.58 & 0.55\\ 

&{OKE 2015 Task 1 example set} & 0.50 & {\color[HTML]{2033FF} \textbf{0.60}} & 0.40 & 0.22 & 0.55 & 0.00 & {\color[HTML]{2033FF} \textbf{0.60}} & 0.55 & 0.00 & 0.50 & {\color[HTML]{2033FF} \textbf{0.60}} & 0.50 & {\color[HTML]{FE0000} \textbf{0.67}} & 0.50\\ 

&{OKE 2015 Task 1 training set} & 0.62 & 0.67 & 0.71 & 0.25 & {\color[HTML]{FE0000} \textbf{0.78}} & 0.00 & 0.61 & {\color[HTML]{FE0000} \textbf{0.78}} & {\color[HTML]{2033FF} \textbf{0.77}} & 0.76 & 0.72 & 0.75 & 0.72 & 0.70\\ 

\midrule
\multirow{6}{*}{\rotatebox{90}{Multilingual}}
&{N$^3$ news.de} & 0.61 & 0.52 & 0.50 & 0.48 & 0.56 & 0.28 & 0.00 & 0.61 & 0.33 & 0.30 & 0.59 & 0.36 & {\color[HTML]{FE0000} \textbf{0.76}} & {\color[HTML]{2033FF} \textbf{0.63}}\\ 

&{Italian Abstracts} & 0.22 & 0.28 & {\color[HTML]{2033FF} \textbf{0.33}} & 0.00 & 0.00 & 0.00 & 0.00 & 0.00 & 0.00 & 0.20 & 0.00 & 0.00 & {\color[HTML]{FE0000} \textbf{0.80}} & {\color[HTML]{FE0000} \textbf{0.80}}\\ 

&{Spanish Abstracts} & 0.25 & 0.33 & 0.26 & 0.00 & 0.24 & 0.27 & 0.00 & 0.47 & 0.00 & 0.31 & 0.33 & 0.31 & {\color[HTML]{FE0000} \textbf{0.75}} &  {\color[HTML]{2033FF} \textbf{0.68}}\\ 

&{Japanese Abstracts} & 0.15 & 0.00 & 0.00 & 0.00 & 0.00 & 0.00 & 0.00 & 0.00 & 0.00 & {\color[HTML]{2033FF} \textbf{0.38}} & 0.00 & 0.00 & {\color[HTML]{FE0000} \textbf{0.54}} & {\color[HTML]{FE0000} \textbf{0.54}} \\ 

&{Dutch Abstracts} & 0.33 & 0.36 & 0.36 & 0.28 & 0.36 & 0.22 & 0.00 & 0.40 & 0.00 & 0.5 & 0.40 & 0.25 & {\color[HTML]{2033FF} \textbf{0.66}} & {\color[HTML]{FE0000} \textbf{0.67}}\\

&{French Abstracts} & 0.00 & 0.00 & {\color[HTML]{2033FF} \textbf{0.28}} & 0.22 & 0.00 & 0.25 & 0.00 & 0.00 & 0.00 & 0.20 & {\color[HTML]{2033FF} \textbf{0.28}} & {\color[HTML]{2033FF} \textbf{0.28}} & {\color[HTML]{FE0000} \textbf{0.80}} & {\color[HTML]{FE0000} \textbf{0.80}}\\ 
\midrule

&\textbf{Average} & 0.45 & 0.48 & 0.50 & 0.36 & 0.55 & 0.24 & 0.11 & 0.53 & 0.31 & {\color[HTML]{2033FF} \textbf{0.59}} & 0.54 & 0.45 & {\color[HTML]{FE0000} \textbf{0.63}} & {\color[HTML]{2033FF} \textbf{0.59}}\\ \midrule

&\textbf{Standard Deviation} & 0.19 & 0.22 & 0.18 & 0.19 & 0.27 & 0.21 & 0.21 & 0.23 & 0.26 & 0.20 & 0.22 & 0.20 & 0.13 & 0.13 \\
\bottomrule
\end{tabular}
\end{table*}

An analysis of our results shows that although the acronym index is an interesting addition for potential improvements, its contribution amounts only to 0.05{\%} F-measure on average over all data sets.
Also, the {popularity} feature improves the results in almost every data set. It can be explained by the analysis of \cite{waitelonis2016don}, which demonstrates that most data sets were created using more popular entities as mentions.
Thus, this bias eases their retrieval\footnote{see the results without popularity using HITS~\url{http://gerbil.aksw.org/gerbil/experiment?id=201701220014}}.
HITS has shown better results on average than Page Rank.\footnote{\url{http://gerbil.aksw.org/gerbil/experiment?id=201701240030}} However, Page Rank did show promising results in some data sets (e.g., Spotlight corpus, AQUAINT, and N3-RSS-500). 

MAG using HITS outperformed the other approaches on 4 of the 17 data sets while achieving comparable results on others, e.g., ACE2004, MSNBC, and OKE data sets. 

Additionally, the performance of MAG can be easily adjusted on each data set by tuning the parameters. For instance, MAG has achieved on \texttt{Micropost2014-test} 0.45 F-measure, but the result may increase around 22{\%} to 0.55 F-measure by disabling the context search.\footnote{\url{http://gerbil.aksw.org/gerbil/experiment?id=201701270035}} 
This improvement is due to the low entity count (1.81 entities per document) and thus missing links among entities inside a document. 

The parameter configuration works not only for huge data sets but also for small data sets or even for disambiguating simple sentences, e.g., ``\texttt{Michael Jordan} is a basketball player and \texttt{Michael Jordan} (born 1957), is an American researcher in Machine learning and Artificial intelligence." If the co-reference resolution is turned on, both ``Michaels" are linked to the same entity (the basketball player), but if this parameter is turned off, MAG is able to find both correctly. 

For the sake of clarity, there is no machine learning in MAG, thus there is no cross-validation set or held-out validation while choosing the parameters.

 
\subsubsection{Multilingual Results}
\label{sec:multilingual}

Here, we show the easy portability and high quality of MAG for many different languages. 
Next to German, Italian, Spanish, French and Dutch, we chose Japanese to show the promising potential of MAG across different language systems. MAG's preprocessing \ac{NLP} techniques are multilingual, thus there is no additional implementation for handling the mentions with different characters. We used the same set of parameters as in the English evaluation but excluded the acronyms as they were only collected for English. Moreover, we performed the Page Rank algorithm over each \ac{KB} in each respective language to collect popularity values of their entities. 
The results displayed in the second part of ~\autoref{engfmeasure} show that MAG, using HITS, outperformed all publicly available state-of-the-art approaches.\footnote{For details see \url{http://faturl.com/multilingualmag/?open}}  Also, Page Rank outperformed HITS score on Dutch. 
The improved performance of MAG is due to its knowledge-base agnostic algorithms and indexing models. 
For instance, although the mention ``Obama" has a high popularity in English, it may have less popularity in Italian or Spanish \ac{KB}s. 
Studies about the generation of proper names support this observation~\cite{dale1995computational,ferreira2017generating}.

\subsubsection{Fine-Grained Evaluation}
\label{sec:fine-grained}
In this analysis, we use an extension of GERBIL done by Waitelonis et al.~\cite{waitelonis2016don}. This extension provides a fine-grained evaluation which measures the quality of a given \ac{EL} for linking different types of entities. This extension also considers the assumption of Van Erp et al.~\cite{van2016evaluating} that a corpus has a tendency to focus strongly on prominent or popular entities which may cause evaluation problems. Hence, the extension evaluates the capability of a given \ac{EL} system for finding entities with different levels of popularity thus revealing its degree of bias towards popular entities.

For this evaluation, we used the same set of parameters as for the English evaluation. The fine-grained analysis shows that MAG is better at linking persons than other types of entities. This can be explained by the indexes created by MAG in the offline phase. They collect last names and rare surfaces for entities. In addition, the results show that MAG is not biased towards linking only popular entities as can be seen in \autoref{tab:finegrained}.\footnote{Detailed information about other data sets can be found here \url{http://gerbil.s16a.org/gerbil/experiment?id=201701260000}}

\begin{table*}[htb!]
\setlength\tabcolsep{2pt}
\small
\centering
\caption{Fine-grained micro F1 evaluation.}
\label{tab:finegrained}
\begin{tabular}{@{} lcccccc @{}}
\toprule
\textbf{Filter }& \textbf{IITB} & \textbf{N3-RSS-500} & \textbf{MSNBC} & \textbf{Spotlight} & \textbf{N3-Reuters-128} & \textbf{OKE 2015} \\
\toprule
Persons & 0.95 & 0.83 & 0.94 & 0.84 & 0.80 & 0.92 \\
Page Rank 10\% & 0.73 & 0.67 & 0.83 & 0.74 & 0.79 & 0.76 \\
Page Rank 10\%-55\% & 0.72 & 0.72 & 0.70 & 0.69 & 0.73 & 0.79 \\
Page Rank 55\%-100\% & 0.73 & 0.71 & 0.73 & 0.75 & 0.76 & 0.82 \\
Hitsscore 10\% & 0.77 & 0.74 & 0.77 & 0.69 & 0.73 & 0.76 \\
Hitsscore 10\%-55\% & 0.69 & 0.66 & 0.64 & 0.69 & 0.79 & 0.78 \\
Hitsscore 55\%-100\% & 0.71 & 0.66 & 0.84 & 0.74 & 0.77 & 0.80
\\
 \bottomrule
\end{tabular}
\end{table*}

\section{Summary}
\label{sec:conclusion}
We presented MAG, a \ac{KB}-agnostic and deterministic approach for multilingual \ac{EL}. 
MAG outperforms the state of the art on all non-English data sets. 
In addition, MAG achieves a performance similar to the state of the art on English data sets. 
Nevertheless, an average of 0.63 F-measure places MAG 1st out of 13 annotation systems. As expected, machine learning-based systems retain their advantages due to their tuned training on the provided data sets. However, we are intrigued by our deterministic and knowledge-base independent performance.
In this paper, we analyzed the influence of different indexing and searching methods as well as the influence of the data set structure in a fine-grained evaluation. We also provided a context search without relying on machine learning as previously done. Moreover, we showed that current ML-based \ac{EL} approaches are strongly biased due to their learned model. This behavior can be seen on multilingual data sets. We also deployed and analyzed the influence of acronyms and last names. To the best of our knowledge, no work has investigated this influence on \ac{EL} and provided a fine-grained evaluation before. 
In the future, we intend to further investigate disambiguation algorithms and MAG's performance in biomedical and earth science domains using the same data set from Wang et. al~\cite{wang2015language}. We also aim to evaluate MAG on different languages such as Arabic using other data sets from~\cite{tsai2016cross} and export the experiment configurations based on the MEX Vocabulary\cite{esteves2015mex} for reproducibility purposes.

\section*{Acknowledgments} 	
This work has been supported by the H2020 project HOBBIT (GA no. 688227) as well as the EuroStars projects DIESEL (no. 01QE1512C) and QAMEL (no. 01QE1549C) and supported by the Brazilian National Council for Scientific and Technological Development (CNPq) (no. 206971/2014-1). This work has also been supported by the German Federal Ministry of Transport and Digital Infrastructure (BMVI) in the projects LIMBO (no. 19F2029I) and OPAL (no. 19F2028A) as well as by the German Federal Ministry of Education and Research (BMBF) within  'KMU-innovativ: Forschung für die zivile Sicherheit' in particular 'Forschung für die zivile Sicherheit' and the project SOLIDE (no. 13N14456).

\bibliographystyle{ACM-Reference-Format}
\bibliography{agdistis}

\end{document}